\pdfoutput=1
\documentclass[11pt]{article}

\usepackage{acl}

\usepackage{times}
\usepackage{latexsym}
\usepackage{hyperref}
\usepackage{booktabs}
\usepackage{graphicx}
\usepackage{amssymb}
\usepackage{xcolor}
\usepackage{forest}
\usepackage{adjustbox}
\usepackage{geometry}
\usepackage{array}
\usepackage{tabularx}
\usepackage{amsmath}
\usepackage{longtable}
\usepackage{multicol}

\usepackage{colortbl}
\usepackage{multirow}
\usepackage{caption}
\usepackage{subcaption}
\usepackage{xspace}
\usepackage[inkscapelatex=false]{svg}
\usepackage{orcidlink}

\usepackage[T1]{fontenc}
\usepackage[utf8]{inputenc}

\usepackage{microtype}

\definecolor{lightblue}{RGB}{173,216,230}
\definecolor{lightpurple}{RGB}{216,191,216}
\newcommand{\corpus}[0]{\texttt{HateDefCon}\xspace}
\newcommand{\defghc}[0]{\texttt{D$_{ghc}$}\xspace}
\newcommand{\defwiki}[0]{\texttt{D$_{wiki}$}\xspace}
\newcommand{\defdict}[0]{\texttt{D$_{dict}$}\xspace}
\newcommand{\deflev}[0]{\texttt{D$_{lev}$}\xspace}
\newcommand{\defisr}[0]{\texttt{D$_{isr}$}\xspace}
\newcommand{\defsyr}[0]{\texttt{D$_{syr}$}\xspace}
\newcommand{\defjor}[0]{\texttt{D$_{jor}$}\xspace}

\newcommand{\redbox}[1]{\cellcolor{red!#1}}
\newcommand{\bluebox}[1]{\cellcolor{blue!#1}}
\newcommand{\greenbox}[1]{\cellcolor{green!#1}}

\newcommand{\KKorcid}{\orcidlink{0000-0002-9349-9554}}
\newcommand{\AMorcid}{\orcidlink{0000-0002-3387-6557}}
\newcommand{\FRorcid}{\orcidlink{0000-0002-1697-8586}}
\newcommand{\ABCorcid}{\orcidlink{0000-0003-4719-3420}}

\newcolumntype{R}{>{\raggedleft\arraybackslash}X}

\title{Untangling Hate Speech Definitions: \\A Semantic Componential Analysis Across Cultures and Domains}

\author{Katerina Korre$^1$\KKorcid \and Arianna Muti$^1$\AMorcid \and Federico Ruggeri$^2$\FRorcid\\ \and \textbf{Alberto Barrón-Cedeño$^1$}\ABCorcid\\
$^1$DIT, University of Bologna, Forlì, Italy \\
   $^2$DISI, University of Bologna, Bologna, Italy\\
    \{aikaterini.korre2, arianna.muti2, federico.ruggeri6, a.barron\}@unibo.it}
\begin{document}
\maketitle
\begin{abstract}
Hate speech relies heavily on cultural influences, leading to varying individual interpretations. 
For that reason, we propose a Semantic Componential Analysis (SCA) framework for a cross-cultural and cross-domain analysis of hate speech definitions. We create the first dataset of hate speech definitions encompassing 493 definitions from more than 100 cultures, drawn from five key domains: online dictionaries, academic research, Wikipedia, legal texts, and online platforms. By decomposing these definitions into semantic components,
our analysis reveals significant variation across definitions, yet many domains borrow definitions from one another without taking into account the target culture. 
We conduct zero-shot model experiments using our proposed dataset, employing 
three popular open-sourced LLMs to understand the impact of different definitions on hate speech detection. 
Our findings indicate that LLMs are sensitive to definitions: responses for hate speech detection change according to the complexity of definitions used in the prompt.

\end{abstract}
\noindent
\textcolor{red}{\textit{\textbf{Warning:} This paper contains offensive language that might be triggering for some individuals.}}

\section{Introduction}
The infeasibility of formulating a universally accepted definition for hate speech and other related concepts (such as toxic language, cyberbullying, and misogyny) is a much discussed topic that permeates not only Natural Language Processing (NLP) research \cite{fortuna-etal-2020-toxic,khurana-etal-2022-hate,pachinger-etal-2023-toward,korre-etal-2023-harmful,nghiem-etal-2024-define} but also expands into the legal and social science fields \cite{Maussen14,Flick_2020,zufall-etal-2022-legal,GuillénNieto+2023}. The lack of a clear definition due to cultural diversity hinders the development of models, as it is unclear what criteria they should be trained to detect.
For instance, consider two definitions, A and B, where only A covers sexual orientation and political opinion criteria.
The statement \textit{``Collectivists are Faggots''} should be labeled as hate speech according to A, and as not hate according to B since B lacks the above-mentioned criteria. 
% Table~\ref{tab:example_definitions} provides an example of a different classification according to definitions with different criteria. 
Cultural perspectives influence how hate speech is perceived; datasets consist of statements produced by individuals within a culture, so the biases reflect, to some extent, the values, norms, and ethics of that culture \cite{bagga-piper-2020-measuring,hershcovich-etal-2022-challenges}. Since most NLP research focuses on English-language data \cite{sogaard-2022-ban}, this cultural dimension is often overlooked, resulting in biases that favor English-speaking cultures.

% \begin{figure}[!t]
%     \centering
%     \includegraphics[width=1\linewidth]{img/hatedefcon.pdf}
%     \caption{\corpus creation pipeline.}
%     \label{fig:dataset_creation}
% \end{figure}

Current NLP approaches are not adequately equipped to address the cultural dependency of hate speech. Existing monolingual hate speech classifiers often lack cultural awareness~\cite{lee-etal-2024-exploring-cross}. Prevailing hate speech taxonomies tend to focus more on legal or academic definitions rather than incorporating cultural dimensions, a gap that can prove detrimental, as hate speech per se and hate speech regulation might influence societal discourse, relationships, and cultural norms, potentially shaping how people interact and express themselves \cite{hietanen}.
\begin{figure}[!t]
    \centering
    \includegraphics[width=1\linewidth]{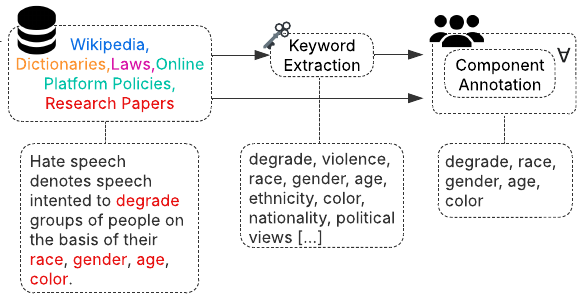}
    %\centerline{\includesvg[width=\columnwidth]{img/pipeline.svg}}
    \caption{\corpus creation pipeline.}
    \label{fig:dataset_creation}
\end{figure}

Inspired by the compositionality principle \cite{compositionality_handbook}, we introduce a component-annotated resource for hate speech definitions, the \corpus dataset. Figure~\ref{fig:dataset_creation} shows our \corpus creation pipeline. We propose a Semantic Componential Analysis (SCA), in which we define a hate speech \textit{definitional component} as a fundamental element or criterion used to define what constitutes hate speech in terms of target, intention/purpose, and act/means. We define \textit{definitional hate speech domains} as the contexts where hate speech definitions emerge. This study focuses on five such domains: legislation, Wikipedia, online dictionaries, research papers, and conduct policies from online platforms or technological companies. In addition, we look at the cultural representation of hate speech. We use \textit{culture}, as the term that encompasses language, ideas, beliefs, customs, codes, institutions, tools, techniques, among other elements.\footnote{\url{https://www.britannica.com/topic/culture}} Distinguishing between culture and language in hate speech definitions is challenging because languages are not confined to a single culture or country. For example, Arabic is spoken across multiple nations with diverse cultural norms, making it difficult to attribute a hate speech definition to a specific cultural context. Additionally, if definitions are collected in a particular language without information on the cultural or national background, it becomes harder to account for variations in meaning, intent, and societal impact across different regions. Unfortunately, most definitions lack any kind of cultural information. We highlight the need for cross-cultural and cross-domain approaches to define hate speech and argue that such definitions should be context-specific, be it cultural, legal, or academic. We advocate for grounding hate speech definitions within these particular domains. This is in line with best practices for tackling subjective tasks \cite{rottger-etal-2022-two}, in which the guideline - or the definition - chosen should consider the downstream use, i.e. the context. 
% , and more specifically, we adopt a novel approach that analyses its most fine-grained semantic aspects. 

In this work, we focus on three research questions: (1) What are the differences among various definitions of hate speech? (2) What is the diversity of these definitions? (3) How do definitions with different components affect the predictions of Large Language Models (LLMs)?
Our contributions are both theoretical and practical. On the theoretical front, our cross-cultural and cross-domain analysis of definitions shows significant variation in components, ranging from broad definitions to highly specific ones. 
Even among the more detailed definitions, which address aspects like the target of hate speech, the intent, and the methods of expressing it, there are differences in their components.
%of those aspects. 
On the practical side, 
% our framework can be easily reproduced to analyze new definitions, particularly when a researcher cannot find an appropriate definition within the \corpus dataset. 
we assess whether LLMs respond better to certain definitions, potentially revealing underlying biases. 
Our results reveal that definitions vary in their components while domains also borrow definitions from one another. 
% In this case, a complete overlap with regard to the components can be hard to achieve. 
When research is culturally specific, borrowing from other domains can be problematic, as it may introduce elements that do not align with the intended cultural context. 
% This is particularly relevant given that many definitions predominantly reflect English-speaking cultures, as seen in dictionaries, research papers, and policies. 
% demonstrate that LLMs are sensitive to the employed hate speech definition.
% Our model experiments demonstrate that the definition affects the LLM response.

The contribution of this paper is threefold:

\begin{itemize}
    \item We propose a semantically informed framework and use it to create \corpus, the first resource for hate speech definitions that amounts to 493 items. 
    \item We conduct a cross-domain and cross-cultural analysis of hate speech definitions and their components.
    \item We assess whether LLMs exhibit a preference for certain definitions, potentially revealing underlying biases.
\end{itemize}

\section{Related Work}
\label{sec:related_work}
To our knowledge, no previous study has approached the analysis of hate speech definitions from the perspective of semantic componential analysis. However, there have been efforts in NLP that focus on decomposing certain concepts into attributes. We review some of these key approaches, while also discussing how the challenge of defining hate speech has been addressed within the NLP community thus far.

\subsection{Decomposition Analysis in NLP}
Some approaches that focus on decomposing concepts into further attributes are \textit{concept analysis}, and \textit{conceptual primitives}. For example, \citet{skuce-meyer-1990-concept} propose using knowledge engineering technology for concept analysis. 
% and introduce CODE (Conceptually Oriented Design Environment), successfully used in a bilingual vocabulary project with and a software documentation project.
More preferred in the field of NLP is Formal Concept Analysis (FCA), which focuses on the relationships between sets of objects and their attributes within a formalized mathematical framework, creating concept lattices to represent these relationships \cite{kamphuis-sarbo-1998-natural}.
FCA is mainly used for ontology construction purposes \cite{li-etal-2005-experiments,moraes-lima-2012-combining,juniarta-etal-2022-organizing}. \citet{pericliev-valdes-perez-1998-procedure-multi}, on a similar note, perform a concept analysis, involving distinguishing classes based on feature values, which is then applied to linguistic tasks. 

% They specifically use Maximally Parsimonious Discrimination (MPD), which infers class profiles to contrast each class from the others, useful in areas like kinship term analysis and phonology. 

At the front of conceptual primitives, i.e. basic units of meaning that cannot be broken down into simpler components within the context of the theory or system they belong to \cite{smith85}, \citet{cambria-etal-2016-senticnet} introduce a resource utilizing hierarchical clustering and dimensionality reduction. 
In their follow-up work, \citet{Cambria_Poria_Hazarika_Kwok_2018} combine sub-symbolic and symbolic AI to automatically discover conceptual primitives from text and link them to commonsense concepts and named entities in a three-level knowledge representation for sentiment analysis.
Similarly, \citet{macbeth2020enhancing} decomposes knowledge and meaning into fundamental perceptual and cognitive structures, improving the interplay between language, expressions, and ontology, and addressing language variation and paraphrasing challenges.

Compared to FCA or conceptual primitives, our SCA framework is a more granular human-driven approach to capturing semantic subtleties at the component level, with particular attention to linguistic context and human interpretability. For example, some domains might emphasize the intent behind hate speech, while others might focus on the effect on the target. With SCA we can pinpoint these differences systematically. In addition, as hate speech is studied across disciplines like law and linguistics, each discipline might have a slightly different understanding of what hate speech is or how it functions. SCA creates a common framework by identifying the shared or different semantic components of hate speech definitions used in these fields, making interdisciplinary research more consistent.
% This enables a more precise semantic breakdown that can potentially complement the mathematical formalisms of FCA or the automatic identification of conceptual primitives.

\subsection{Hate Speech and the Issue of Definitions}
Recent NLP work has focused on comparing harmful language definitions (abusiveness, toxicity, offensiveness, etc.) rather than comparing hate speech definitions per se. 
\citet{fortuna-etal-2020-toxic} focus on clarifying applied categories and homogenizing different datasets by treating class representations as FastText vectors rather than relying on their definitions. 
% Their work involves two main experiments. 
% The first experiment compares categories across annotated datasets to clarify and homogenize the applied categories. They calculate the average centroid for each message in a category and measure category homogeneity using cosine similarity. The second experiment uses the Perspective API to evaluate the classifier's ability to distinguish standardized categories from non-harmful messages through binary classification. 
While \citet{fortuna-etal-2020-toxic} compare terms that are close semantically, such as toxicity, abusiveness, and aggressiveness, our work focuses exclusively on hate speech definitions.
% As their final remark, the authors stress the importance of providing guidelines for annotation schemas and avoiding the creation of new categories unless absolutely necessary, with clear examples and justifications when new categories are introduced. 
\citet{pachinger-etal-2023-toward} review definitions of uncivil, offensive, and toxic comments across 23 papers from various fields, aiming to foster unified scientific resources. 
Their work highlights the need for consistent terminology across disciplines to enhance clarity and application in scientific research.
We embrace this mindset presented in both \citet{fortuna-etal-2020-toxic} and \citet{pachinger-etal-2023-toward} by offering a resource that includes a wide range of hate speech definitions ---from general to specific, and from various sources--- allowing researchers to select from established, culturally relevant options.
This helps to avoid the creation of custom definitions when unnecessary, which could introduce bias into experimental models, beginning with the annotation process.

The work that is more similar to ours is the one by \citet{khurana-etal-2022-hate}, who develop hate speech criteria with input from legal and social science perspectives to help researchers create precise definitions and annotation guidelines. 
They propose five criteria: target groups, dominance, perpetrator characteristics, type of negative group reference, and potential consequences/effects. 
Rather than prescribing a single definition, they offer a meta-prescriptive, modular approach, allowing for adjustments based on specific tasks. 
This approach emphasizes the role of subjectivity in definitions and addresses the issues arising from varying and vague definitions, which can lead to inconsistencies and problematic expectations about dataset annotations.
We build upon the work of \citet{khurana-etal-2022-hate} by introducing a more fine-grained framework, and extend the study by examining the impact of the components when applying different definitions when prompting LLMs.

%  \citet{lee-etal-2024-exploring-cross} select posts from the SBIC dataset \cite{sap-etal-2020-social}, which mainly represents North America, 
% and gather posts from four geographically diverse English-speaking countries —Australia, the United Kingdom, Singapore, and South Africa— using culturally specific hateful keywords identified through a survey. Annotations are collected from these four countries, along with the United States, to create representative labels for each. Their analysis reveals statistically significant differences in hate speech annotations across the countries. On a more theoretical front, \citet{hietanen} explore the challenge of defining hate speech, reviewing legal and academic approaches, and ultimately proposing four categories: (A) teleological, focusing on intent, (B) consequentialist, based on outcomes, (C) formal, centered on the nature of the speech act, and (D) relativist, shaped by societal consensus. They briefly touch on the cultural dimension, suggesting that speech regulation can subtly influence culture, discourse, and relationships.

\section{Semantic Componential Analysis} \label{sec:sca}

Semantic Componential Analysis (SCA) is a linguistic technique used to break down the meanings of words or phrases into their constituent parts or features. 
This method, central to structural linguistics, has been in use since the 1950s \cite{Lounsbury,goodenough,Nida1975-NIDCAO}. 
It is %a method 
based on the \textit{principle of compositionality}, which states that the meaning of a complex expression is derived from the meanings of its components and the rules governing their combination \cite{compositionality_handbook}. 
SCA involves compiling a detailed list of specific examples for each term within a group of contrasting terms.
Each example is described using a set of relevant attribute dimensions \cite{KRONENFELD2005361}. 
SCA primarily examines words through organized sets of semantic features, which are marked as `present' or `absent', using +/- symbols \cite{GEERAERTS2006709}. 
% An example is illustrated in Table~\ref{tab:kinship}. 
Appendix~\ref{sec:appendix:sca-example} shows an example.

% \begin{table}[!t]
%     \centering
%     \resizebox{\columnwidth}{!}{
%     \begin{tabular}{lccccc}
%         \toprule
%            & Parent & Sibling & Male & Female & \\
%         \midrule
%         Father    & + & - & + & - \\
%         Mother    & + & - & - & +  \\
%         Brother   & - & + & + & -  \\
%         Sister    & - & + & - & +  \\
%        \bottomrule
%     \end{tabular}
%     }
%     \caption{Kinship domain examples characterized by attribute dimensions. The domain is kinship terms, with attribute dimensions including Parent, Sibling, Male, Female. Each example is marked with '+' if the feature is present, '-' if absent.}
%     \label{tab:kinship}
% \end{table}

SCA is an approach that enables us not only to break down terms into their individual components, but also to explore various types of meanings as categorized by \citet{leech1990semantics}. 
By comparing terms side by side, we can enhance our analysis, going beyond the conceptual meaning (the stable meaning across contexts), touching upon the connotative and social meanings. 
These latter meanings can vary depending on cultural and social contexts. 

\citet{KRONENFELD2005361} describes the identification of components. 
%Analysts typically use experimental methods. 
One approach involves systematically alternating attributes to determine which distinctions are essential for differentiating terms. 
Another method is to gather descriptions from informants about differences between terms or subsets of terms, gradually building a set of potential semantic components based on these descriptions. The methodology presented in this paper draws inspiration from the latter approach.
% Originally, it was believed that attribute dimensions had to be drawn from a universal, predefined set (an `etic grid'), but it's now recognized that attributes can come from any source as long as they are well-defined and relevant to the domain under study.

\section{The \corpus Dataset} 
\label{sec:hatedefcon_creation}

\begin{figure*}[!t]
    \centering
    \includegraphics[width=\linewidth]{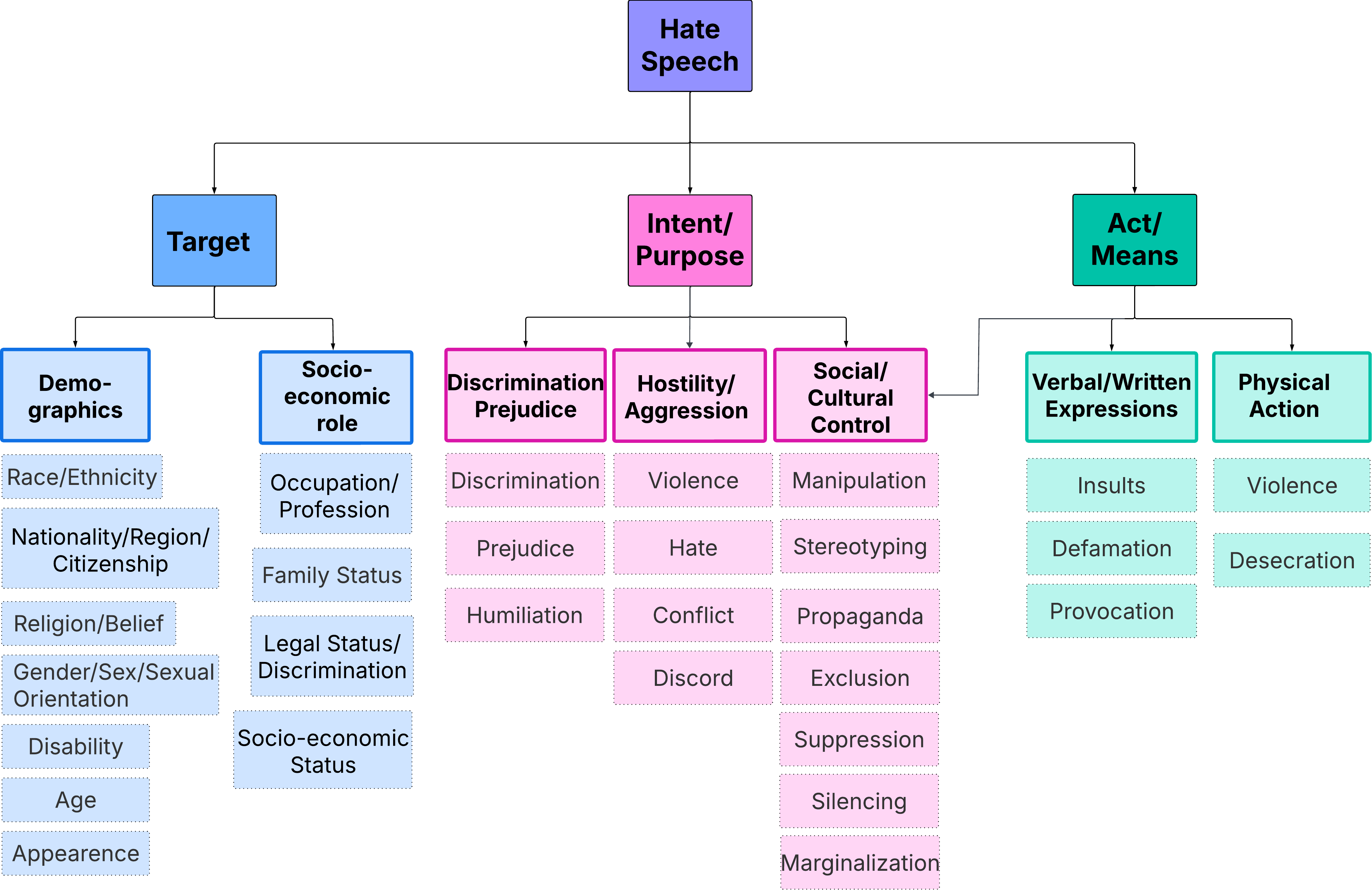}
    \caption{SCA component hierarchy in \corpus.}
    \label{fig:hierarchy_graph}
\end{figure*}

We analyze and compare hate speech definitions via SCA to identify potential cross-domain and cross-cultural differences.
We collect 493 definitions from five different domains in which we can find hate speech definitions (\S\ref{subsec:data}). 
% i.e., hate speech legislation, hate speech articles from Wikipedia, dictionary definitions, definitions found in NLP research papers, online conduct policies of online platforms. 
We provide information about the respective sources and a final corpus analysis to gain cross-domain and cross-cultural insights (\S\ref{subsec:component_annotation}).
Figure~\ref{fig:dataset_creation} summarizes the \corpus dataset creation pipeline. We collect definitions from the selected domains, we extract keywords which we use to annotate as components for each definition. \corpus, along with all the results of the experiments is publicly available.\footnote{\url{https://github.com/katkorre/SCA-of-Hate-Speech.git}}

\subsection{Data}\label{subsec:data}

\paragraph{Hate Speech Laws.}
% The first context in which we examine the definition of hate speech is hate speech laws. 
We source definitions from the Global Handbook of Hate Speech Laws website,\footnote{\url{https://futurefreespeech.org/global-handbook-on-hate-speech-laws/}.}
% Our data were sourced from the Global Handbook on Hate Speech Laws website\footnote{\href{https://futurefreespeech.org/global-handbook-on-hate-speech-laws/}{Global Handbook of Hate Speech Laws}}, 
a comprehensive resource that provides access to existing hate speech legislation from countries worldwide, including the United Nations and European Union levels.
The legislations are already available in English.
We exclude countries for which their legislative texts were not available. The full list of the countries is available in Appendix~\ref{app:laws_list}.
% It is important to note that some countries only provided metadata without legislative texts. Therefore, these countries were excluded from our experiments.

\paragraph{Wikipedia.}
We scrape Wikipedia articles in different languages and manually extract the relevant portions of the definitions. 
We automatically translate extracted portions into English.\footnote{\url{https://www.deepl.com/en/translator}.}

\paragraph{Dictionaries.}
% For dictionary definitions, 
We consult several dictionaries across multiple languages. 
As with the Wikipedia definitions, we translate the extracted content into English using machine translation.

\paragraph{NLP Research Papers.}
% We extracted definitions from NLP research papers. 
We use the Hate Speech Dataset Catalogue \footnote{\url{https://hatespeechdata.com/}} to navigate through research articles and datasets, focusing specifically on definitions of hate speech.
We do not include other related terms like toxicity or abusive language, as our primary focus is hate speech. 
% This emphasis is due to the fact that hate speech carries significant historical and social implications, and the legislation primarily targets this specific term.

\paragraph{Online Conduct Policies.}
We use the conduct policies from online platforms or technology companies that address the use of hate speech or harmful language online.\footnote{The conduct policies were collected from X, Meta, Microsoft, Pinterest, Snapchat, YouTube, Reddit, TikTok, and Discord.}

% \subsubsection{NLP methods}
% For the exploration of the laws we used similarity and clustering methods. First of all, we used the Universal Sentence Encoders by Tensorflow to calculate the semantic similarity of the sentences/definitions within the different hate speech definitional contexts. 
% Secondly, we used TF-IDF to extract information about keywords and semantice properties that will help us later cluster the contexts and their contents. 

\subsection{Component Annotation}\label{subsec:component_annotation}

We engage three annotators for SCA labelling.
% For the purposes of annotation and subsequent componential analysis, we engaged three annotators. 
Annotators are proficient in English, have experience in annotating tasks, and are familiar with hate speech research. All three of them annotate the entire corpus.
We provide annotators with detailed instructions on annotating definitions following our framework, as outlined in Appendix~\ref{app:annotation_guidelines}.
% Our approach to component annotation follows the method described by \citet{KRONENFELD2005361}. 
To compile an initial set of components, we employ keyword extraction and $tf$-$idf$ on the collected definitions. A manual review reveals that these automated methods fail to capture some components, highlighting the necessity for human annotation. Despite this limitation, these computational techniques offer a valuable starting point. Using the resulting list, annotators then perform binary annotations, indicating the presence or absence of each component~\cite{KRONENFELD2005361}. 

To ensure objectivity in the annotation process and a very fine-grained representation of the components, we establish that a definition contains a given component if a derivative of the component's word appears in the annotation. For instance, if the word `abusive' is present in the definition, the annotator marks the component `abuse' as present. 
A challenge we encounter during annotation is the treatment of synonyms. 
% One challenge we encountered during annotation was how to handle synonyms. 
For instance, the words `harm' and `hurt' can be considered synonymous in the context of hate speech definitions, but they are not derivatives of the same root word.\footnote{`Harm' comes from Middle English 'harm', while 'hurt' comes from Middle English `hurten'}
To maintain consistency and avoid subjective interpretation, which could lead to different grouping of the synonyms by the annotators, we only annotate derivatives of the target term (e.g., marking `harm' only if words like `harmful' or `harmed' appear). 
% This approach ensures that annotators do not mark synonyms based on their personal conception, but rather rely on the presence of directly related word forms. 
Additionally, missing components are labeled as `undefined component'.
This process allows us to comprehensively gather all components deemed crucial by the annotators. 

\begin{figure}[!t]
    \centering
    \includegraphics[width=1\linewidth]{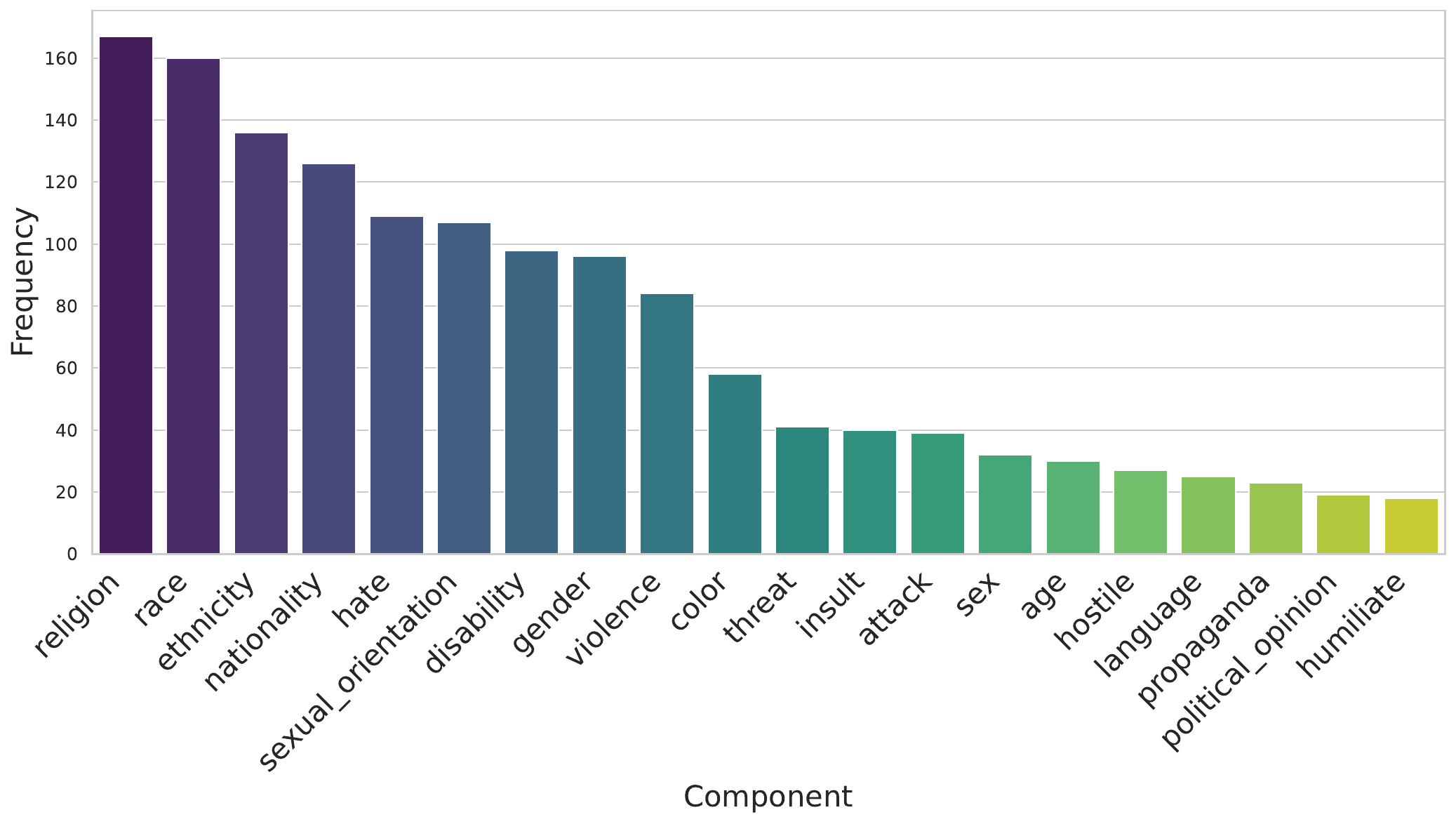}
    \caption{Top 20 most frequent components across all definitions.}
    \label{fig:app:overall_frequencies}
\end{figure}

To finalize the dataset, we keep all annotated components, as a manual inspection showed that most of the disagreement derived from the fact that one of the annotators missed a component. The average inter-annotator agreement (IAA) measured via Cohen's kappa~\cite{mchugh2012interrater} is 0.64 (see Appendix~\ref{app:iaa} for more details). 
% The pairwise IAA can be found in the in Table~\ref{tab:iaa}. 
Figure~\ref{fig:hierarchy_graph} reports the component hierarchy resulting from the annotation study (see Appendix~\ref{app:comp_hierarchy} for a fine-grained version of the hierarchy).
Figure  \ref{fig:app:overall_frequencies} reports the overall frequency for the top 20 components. The most frequent target components are those mostly related to racism: religion, race, ethnicity and nationality. This indicates a stronger emphasis on aspects of identity that hold deep historical and social significance in discriminatory narratives, compared to biological attributes such as age and sex, which, while still present, appear less prominently.

% The hierarchy of the components can be found in Figure~\ref{fig:hierarchy_graph}. 
% An even more fine-grained version of the graph can be found in the Appendix~\ref{app:comp_hierarchy}, in which the specific keyword components under each final node are also included.

\section{Case Study: Definition Comparison}\label{case_study}

\begin{table*}[!t]
\centering
\small
\begin{tabularx}{\textwidth}{lc p{4.5cm} p{5.7cm}}  
\toprule
\textbf{Domain} & \multicolumn{1}{c}{\textbf{No. Definitions}} & \multicolumn{1}{c}{\textbf{Culture Distribution}} & \multicolumn{1}{c}{\textbf{Components}} \\
\midrule
Law & 116 & All cultures appear once & Race, Religion, Hate, Nationality, Ethnicity \\
Wikipedia & \,\,49 & All cultures appear once & Religion, Gender, Sexual Orientation, Race, Disability \\
Research Paper & \,\,29 & English (13), German (3), Arabic~(2), Indonesian (2), the rest appear once & Gender, Religion, Sexual Orientation, Ethnicity, Race \\
Dictionary & \,\,21 & English (7), Italian (5), the rest appear once & Attack, Gender, Sexual Orientation, Religion, Hate \\
Platform & \,\,\,\,278 & All cultures appear once per platform & Disability, Ethnicity, Gender, Nationality, Race \\
\addlinespace
\midrule
\textbf{Total} & 493 & & \\
\bottomrule
\end{tabularx}
\caption{Number of definitions, culture distribution, and top 5 components per domain.}
\label{tab:data_summary_combined}
\end{table*}

% \begin{table}[!t]
% \centering
% \small
% \begin{tabularx}{\columnwidth}{Xr p{3.5cm}}  
% \toprule
% \textbf{Domain} & \multicolumn{1}{c}{\textbf{No. Instances}} & \multicolumn{1}{c}{\textbf{Culture Distribution}} \\
% \midrule
% Law & 116 & All cultures appear once \\
% \hline
% Wikipedia & 49 & All cultures appear once \\
% \hline
% Research Paper & 29 & English (13), German (3), Arabic (2), Indonesian (2), the rest appear once \\
% \hline
% Dictionary & 21 & English (7), Italian (5), the rest appear once \\
% \hline
% Platform & 9 & All policies are in English \\
% \addlinespace
% \midrule
% \textbf{Total} & 224 &  \\ 
% \bottomrule
% \end{tabularx}
% \caption{Corpus domain and culture distribution.}
% \label{tab:data_summary}
% \end{table}

\paragraph{Qualitative Analysis.} We perform
a qualitative assessment of the definitions and provide some initial insights into our data. 
Our observations reveal that many definitional domains, particularly in research papers and Wikipedia articles, frequently borrow definitions from other sources. 
For instance, 17 out of 30 research papers reference definitions from other academic papers, legislation, or platform policies. Wikipedia, on the other hand, often relies on definitions from the Cambridge Dictionary. 
Through translation, definitions might reflect the culture of the source text, and in collaborative projects, cultural elements from multiple backgrounds may become blended.
\paragraph{Distributions.} Table~\ref{tab:data_summary_combined} reports culture distribution for each domain in \corpus.
A great disparity is evident with the English definitions, which appear most often in research papers and dictionaries, while platform policies are always the same text but translated in different languages. Definitions from other cultures appear fewer than 5 times, and most occur only once in the dataset.
Table~\ref{tab:data_summary_combined} also reports the 5 most frequent components per domain.
The most common components are related to the target of hate speech, mainly being associated with religion, ethnicity, and gender.
% 
% The most common components are related to the target, particularly concepts associated with religion, ethnicity and gender. 
% With regard to the intent/purpose we see `hate' and `attack'. Figure~\ref{fig:app:overall_frequencies} in Appendix also displays the top 20 most frequent frequent components across all definitions, irrespective of domain or culture. 

% \begin{table}[!tb]
% \small
% \centering
% \begin{tabular}{p{2.5cm}p{4cm}}
% \toprule
% \textbf{Domain} & \textbf{Components} \\ \hline
% Law                 & Race, Religion, Hate, \\
%                      & Nationality, Ethnicity \\ \hline
% Wikipedia            & Religion, Gender, Sexual \\
%                      & Orientation, Race, Disability \\ \hline
% Research paper      & Gender, Religion, Sexual \\
%                      & Orientation, Ethnicity, Race \\ \hline
% Dictionary        & Attack, Gender, Sexual \\ 
%                      & Orientation, Religion, Hate \\ \hline

% Platform             & Disability, Ethnicity, \\
%                      & Gender, Nationality, Race \\ \bottomrule

% \end{tabular}
% \caption{Top 5 components per domain.}
% \label{tab:top_5_domain_components}
% \end{table}
% % 

\begin{table*}[!t]
\centering
\small
\resizebox{2.1\columnwidth}{!}{
\begin{tabular}{>{\raggedright\arraybackslash}p{2.5cm}>{\raggedright\arraybackslash}p{3.cm}>{\raggedright\arraybackslash}p{3.5cm}>{\raggedright\arraybackslash}p{2.5cm}>{\raggedright\arraybackslash}p{3.5cm}}
\toprule
\textbf{Components} & \textbf{\citet{mulki-etal-2019-l}} & \textbf{Israel Legislation} & \textbf{Syria Legislation} & \textbf{Jordan Legislation} \\
\midrule
\textbf{Target} & & & & \\
\quad \textbf{Demographics} & Ethnicity, Religion, Sexual Orientation, Color, Gender, Nationality, Race & Ethnicity, Religion, Race, Color & Race & Ethnicity, Religion, Race \\
\midrule
\textbf{Intent} & & & & \\
\quad \textbf{Discrimination} & Disparage &  Humiliation, Degradation & Race & Prejudice \\
\quad \textbf{Hostility} & N/A & Hostility, Enmity & N/A & Hurt, Hate, Conflict, Violence \\
\quad \textbf{Cultural Control} & N/A & Persecution, Degradation & N/A & N/A \\
\midrule
\textbf{Act} & & & & \\
\quad \textbf{Expression} & Disparage & Hostility, Humiliation, Enmity & N/A & Hate \\

\quad \textbf{Physical Action} & N/A & N/A & N/A & Hurt, Conflict, Violence, Terrorism \\
\quad \textbf{Manipulation} & N/A & Persecution, Cultural Control & N/A & N/A\\ \bottomrule
\end{tabular}%
}
\caption{Comparative analysis of components of \citet{mulki-etal-2019-l} vs. the legislation of countries where Levantine Arabic is spoken in.}
\label{tab:comparison}
\end{table*}

\paragraph{Cross-Cultural Case Study.} We employ SCA to describe potential inconsistencies between collected hate speech definitions and the cultural reality through a real case study.
\citet{mulki-etal-2019-l} develop a Levantine Hate Speech and Abusive Twitter dataset in an attempt to bring into the spotlight less-spoken Arabic varieties. 
However, in their paper, the authors refer to the hate speech definition by \citet{nockleby2000hate}, which defines hate speech as ``any communication that disparages a person or a group on the basis of some characteristic such as race, color, ethnicity, gender, sexual orientation, nationality, religion, or other characteristic''. 
Levantine Arabic is spoken in many Middle Eastern countries, such as Syria, Jordan, and Israel. 
Our SCA framework allows us to see if the definitions available for these countries overlap with the one used in \citet{mulki-etal-2019-l} (Table~\ref{tab:comparison}). 
While they all differ in the intent and act, some targets overlap, although the definition from \citet{mulki-etal-2019-l} is more comprehensive, considering also sexual orientation and gender, which are not taken into account by the other definitions. Israel's legislation seems to have the most detailed provisions, including concepts of hostility and cultural control, whereas Syria's legislation is narrowly focused on race, and Jordan's legislation emphasizes physical action and terrorism. This variation can affect annotation and prompting, resulting in varied interpretations of hate-related content. Therefore, tailoring prompts to align with specific regional definitions is essential for achieving consistent model behavior for a specific culture.

\section{LLM Sensitivity to Definitions}
\label{sec:model_experiments}
We evaluate the ability of LLMs to perform the task of binary hate speech classification based on the definitions provided. 
% In particular, our objective is twofold: expose the biases of LLMs in what they consider to be hate speech, and 
We assess how much LLMs rely on the definitions of hate speech provided rather than their own internal knowledge.
% By doing so, we expose the biases of LLMs in what they consider to be hate speech. 
% In addition, we assess how much LLMs rely on the definition provided by us rather than on their internal knowledge. 

\paragraph{Models and Data.}
We experiment with three open-source popular state-of-the-art LLMs: Llama3\footnote{\texttt{Llama3-8B-instruct.}}, Mistral2\footnote{\texttt{Mistral-7B-Instruct-v0.2}.}, and Phi3-mini.\footnote{\texttt{Phi-3-mini-4k-instruct}.}
We consider the Gab Hate Corpus (GHC) on online hate speech conversations~\cite{kennedy_introducing_2022}.
GHC consists of 27,665 posts from the social network service gab.com and is annotated for the presence of ``hate-based rhetoric'' by a minimum of three annotators. The posts are annotated based on a coding typology created by synthesizing definitions of hate speech from legal precedents, existing hate speech coding frameworks, and insights from psychology and sociology. This typology includes hierarchical labels that denote dehumanizing and violent speech, as well as indicators related to targeted groups and rhetorical framing. This dataset captures various dimensions of hate speech, making it well-suited for a detailed and fine-grained analysis and testing with multiple definitions.
Although GHC is annotated in a multi-class fashion, we consider a binary setting: hate vs not hate.
To limit computational overhead, we select 500 instances from the corpus, equally divided between the two classes.

% 
% \footnote{\url{https://huggingface.co/meta-llama/Meta-Llama-3-8B}}, \texttt{Mistral-7B-v02}\footnote{\url{https://huggingface.co/mistralai/Mistral-7B-Instruct-v0.2}}, and \texttt{Phi-3-mini-4k}.\footnote{\url{https://huggingface.co/microsoft/Phi-3-mini-4k-instruct}} 
% We select the Instruct-tuned version for all models.
% Llama3 features several improvements over preceding versions, such as a better tokenizer with a vocabulary of 128$k$ tokens and training on 15$T$ tokens. 
% Mistral2 is a fine-tuned version of Llama2 using group-query attention. 
% In particular, the 7B version has been reported to outperform competitors like Llama2-7B and Llama2-13B. 
% Phi3-mini is a 3.8B parameter model that supports input contexts up to 4k tokens. 
% While comparably smaller than other popular LLMs, this model has shown remarkable performance in reasoning and language understanding benchmarks.

% \paragraph{Metrics.}
% We follow standard practice for binary classification in hate speech and compute the F$_1$ score on the positive class~\cite{bose-su-2022-deep}.

\paragraph{Setup.}
We perform our experiments in a zero-shot setting to understand the impact of different definitions (i.e. different components) on model performance without any additional fine-tuning or prompting strategies. 
To prompt the models, we use three definitions of hate speech. We select the definitions deliberately after manual inspection, aiming to assess whether the varying degrees of component coverage influence the prompting outcomes. This approach includes one definition with medium component coverage, a highly detailed one, and a very general one.
We use three definitions: \defghc comes from the GHC dataset, it includes several components but is not the most comprehensive one; 
\defwiki is the definition provided by the Macedonian Wikipedia page, it offers a more detailed and varied perspective; and \defdict is a general definition from the Merriam-Webster dictionary (see Appendix~\ref{app:ghc-definitions} for more details).
We set the temperature to zero to exclude variations in the generated responses.

\section{Results}
\label{sec:results}
Table \ref{tab:results} summarizes the results of the classification task according to the three definitions.
% on a balanced sample of 500 instances.
Although Llama3 performs consistently better than other models, the complexity of the definitions affects the outputs of the three models differently.
In Llama3, the comprehensiveness of the definition is directly proportional to the performance: the more comprehensive the definition is, the more the performance increases; in Mistral2, we observe a reverse tendency. 
Phi3-mini outperforms Mistral2 in all settings, although being half the size of Mistral2. 
Additionally, in Phi3-mini, we observe the smallest variety of responses based on the definitions, with \defghc performing the best.

There are cases in which Llama3 refuses to answer
% , due to the safety fine-tuning that has undergone, 
for safeguarding the generation process over harmful content.
% However, after manual inspection, it is still not clear why the safeguard activates only in certain instances. 
This tendency is also proportional to the completeness of the definition. The less comprehensive the definition, the more the model refuses to answer.
For \defwiki, the model refuses to answer to 1\% of times, 1.6\% for \defghc, and 5\% for \defdict. 
% \defghc refuses to answer 1.6\% of times, \defwiki 1\% and \defdict 5\%.

\begin{table}[!t]
\small
    \centering
    \begin{tabularx}{\linewidth}{Xccc} 
        \toprule
        \textbf{Model} & \defghc & \defwiki & \defdict  \\
        \midrule
         Llama3 & 0.67 & 0.78 & 0.70 \\
         Mistral2 & 0.56 & 0.53 & 0.58 \\
         Phi3-mini & 0.60 & 0.58 & 0.58 \\
        \bottomrule
    \end{tabularx}
    \caption{Classification performance (F$_1$score) using difference hate speech definitions.}
    \label{tab:results}
\end{table}

% There are two reasons why this happens: either because the definition is too narrow, or because it contains slurs. Llama3 has undergone safety fine-tuning for safeguarding the generation process over harmful content, however it is not clear why the safety parameter activates only in certain instances.

\paragraph{Error Analysis.}
We conduct an error analysis of the answers generated by Llama3.
Our error analysis encompasses two aspects: how the model perform considering the definitions against the gold labels, and how the predictions differ according to the definition used.
% 
% \begin{figure}
%     \centering
%     \includegraphics[width=1\linewidth]{img/matrix.png}
%     \caption{Confusion matrix with LlaMa3}
%     \label{fig:matrix}
% \end{figure}
% 
Figure~\ref{tab:matrices_general_defs} shows the confusion matrices with respect to the three definitions. 
In all three settings, the false negatives can partly be attributed to the lack of components in the definition. 
For instance, \defghc focuses on violence and hate, excluding all other forms of hate speech. 
% The application of the definitions results in similar false negative rates—60, 60, and 57 out of 500 instances—indicating that these laws share common gaps in their definitions.
As a result, discrimination against a group of people that does not exhibit violence or hate often goes undetected. 
% Consider the statement: \textit{``Illegal aliens receive preferential treatment from state and federal judges, as well as from state and federal prosecutors, compared to US citizens''}, which implies negative bias towards undocumented immigrants by reinforcing stereotypes. 
However, the presence or absence of such components is not a definitive factor. 
Take the following example sentence: \textit{``Real friends don't let their friends get infected with communism''}.
The sentence is incorrectly categorized as non-hateful by \defwiki, even though it contains political views as a component. 
While not conclusive, analyzing the components of misclassified examples can contribute to understanding the reasons behind the classifications.

\begin{figure}[!t]
\tiny
\centering

% First Row (Two subtables side by side)
\begin{minipage}{0.45\linewidth}
\centering
\begin{subtable}[t]{1\textwidth}
\centering
\setlength{\arrayrulewidth}{0.3mm}
\setlength{\tabcolsep}{5pt} % Reduce padding to make tables smaller
\renewcommand{\arraystretch}{1.2}
\begin{tabular}{|c|c|c|}
\hline
\textbf{True \textbackslash Predicted} & \textbf{0} & \textbf{1} \\
\hline
\textbf{0} & \redbox{70} 226 & \redbox{10} 24 \\
\hline
\textbf{1} & \redbox{30} 103 & \redbox{50} 147 \\
\hline
\end{tabular}
\caption{\defghc}
\end{subtable}
\end{minipage}%
\hspace{2em} % Horizontal space between subtables
\begin{minipage}{0.45\linewidth}
\centering
\begin{subtable}[t]{1\textwidth}
\centering
\setlength{\arrayrulewidth}{0.3mm}
\setlength{\tabcolsep}{5pt} 
\renewcommand{\arraystretch}{1.2}
\begin{tabular}{|c|c|c|}
\hline
\textbf{True \textbackslash Predicted} & \textbf{0} & \textbf{1} \\
\hline
\textbf{0} & \bluebox{65} 214 & \bluebox{15} 36 \\
\hline
\textbf{1} & \bluebox{20} 67 & \bluebox{55} 183 \\
\hline
\end{tabular}
\caption{\defwiki}
\end{subtable}
\end{minipage}
\vspace{1.5em} % Vertical space between rows

% Second Row (Two subtables side by side)
\begin{minipage}{0.45\linewidth}
\centering
\begin{subtable}[t]{1\textwidth}
\centering
\setlength{\arrayrulewidth}{0.3mm}
\setlength{\tabcolsep}{5pt} % Reduce padding to make tables smaller
\renewcommand{\arraystretch}{1.2}
\begin{tabular}{|c|c|c|}
\hline
\textbf{True \textbackslash Predicted} & \textbf{0} & \textbf{1} \\
\hline
\textbf{0} & \greenbox{68} 222 & \greenbox{12} 28 \\
\hline
\textbf{1} & \greenbox{30} 102 & \greenbox{50} 148 \\
\hline
\end{tabular}
\caption{\defdict}
\end{subtable}
\end{minipage}%

% \bigskip
% \begin{subtable}[t]{0.28\textwidth}
% \centering
% \setlength{\arrayrulewidth}{0.3mm}
% \setlength{\tabcolsep}{5pt} % Reduce padding to make tables smaller
% \renewcommand{\arraystretch}{1.2}
% \begin{tabular}{|c|c|c|}
% \hline
% \textbf{True \textbackslash Predicted} & \textbf{0} & \textbf{1} \\
% \hline
% \textbf{0} & \redbox{70} 226 & \redbox{10} 24 \\
% \hline
% \textbf{1} & \redbox{30} 103 & \redbox{50} 147 \\
% \hline
% \end{tabular}
% \caption{\defghc}
% \end{subtable}%
% \vspace{1.5em} % Adds space between subtables
% \begin{subtable}[t]{0.28\textwidth}
% \centering
% \setlength{\arrayrulewidth}{0.3mm}
% \setlength{\tabcolsep}{5pt} 
% \renewcommand{\arraystretch}{1.2}
% \begin{tabular}{|c|c|c|}
% \hline
% \textbf{True \textbackslash Predicted} & \textbf{0} & \textbf{1} \\
% \hline
% \textbf{0} & \bluebox{65} 214 & \bluebox{15} 36 \\
% \hline
% \textbf{1} & \bluebox{20} 67 & \bluebox{55} 183 \\
% \hline
% \end{tabular}
% \caption{\defwiki}
% \end{subtable}%
% \hspace{1.5em} % Adds space between subtables
% \begin{subtable}[t]{0.28\textwidth}
% \centering
% \setlength{\arrayrulewidth}{0.3mm}
% \setlength{\tabcolsep}{5pt} % Reduce padding to make tables smaller
% \renewcommand{\arraystretch}{1.2}
% \begin{tabular}{|c|c|c|}
% \hline
% \textbf{True \textbackslash Predicted} & \textbf{0} & \textbf{1} \\
% \hline
% \textbf{0} & \greenbox{68} 222 & \greenbox{12} 28 \\
% \hline
% \textbf{1} & \greenbox{30} 102 & \greenbox{50} 148 \\
% \hline
% \end{tabular}
% \caption{\defdict}
% \end{subtable}

\caption{LLama3 confusion matrices with respect to each hate speech definition. When the model refuses to answer, we set the predicted label to zero.}\label{tab:matrices_general_defs}
\end{figure}

\paragraph{How do definitions affect the output?}
We extract all instances for which the answer of Llama3 varies according to the definition, regardless of the ground truth. 
\defghc and \defwiki differ in 54 instances, while \defdict differs from \defghc and \defwiki in 48 instances each.
% 
% \defwiki and \defdict and \defghc and \defdict both differ in 48 instances.
We observe that when the model is prompted with \defwiki, it tends to identify more personal attacks, which are often discarded in \defghc and \defdict as false negatives since such definitions identify group of people as targets of hate speech, rather than individuals. 
Moreover, with \defwiki, LLMs tend to identify more instances that are hateful with respect to political views and undocumented migrants. 
Indeed, the majority of instances that are correctly identified by \defwiki and misclassified by \defghc and \defdict, contain terms like liberals, communists, and illegal aliens with a negative connotation. In short, we see that the models indeed carry their own biases which are evident in how they classify instances even if we change the definition in the prompt, indicating that they rely more on their own internal knowledge than solely on the definition. 

\begin{table}[!t]
    \centering
    \small
    \begin{tabularx}{\linewidth}{l|RRRR} % 'X' columns will automatically adjust width
        \toprule
        \bf Def. & \deflev & \defisr & \defsyr & \defjor \\
        \midrule
        \deflev     & 1.00      & 0.74       & 0.59       &  0.50     \\
        \defisr & - & 1.00 & \textbf{0.75} & 0.67 \\
        \defsyr  & - & - & 1.00 & 0.50 \\
        \bottomrule
    \end{tabularx}
    \caption{Agreement across definitions via F$_1$ score. 
    % on the positive classs that represents the agreement between the models when they use different definitions.
    }
    \label{tab:results_cross}
\end{table}

\paragraph{Cross-Cultural Analysis.}
We go back to the case study of the Levantine dataset (Table~\ref{tab:comparison}) to explore how these definitions affect prompting and whether cultural biases may arise as a consequence.
% Considering the variations in definitions, we go back to the case study of the Levantine dataset presented in Table~\ref{tab:comparison}, and we explore how these definitions affect prompting and whether cultural biases may arise as a consequence.
We compare predictions for the definition from the original Levantine dataset \cite{mulki-etal-2019-l} (\deflev), the one for Syria (\defsyr), the one for Israel (\defisr), and the one for Jordan (\defjor).
The four definitions are in Appendix~\ref{app:prompts}.
We consider Llama3 for our analysis as our best-performing model. 
% We perform our experiments with our best model, i.e., \texttt{llama3-8b-Instruct}. 
% Table \ref{tab:results_comp} shows the F$_1$ score on the positive class.
% We observe that 
Llama3 achieves the top performance (F$_1$=0.74) with \defisr and \defsyr, exceeding most of the results with the original definition.
This is an interesting result since \defsyr is the least comprehensive definition, with only one target component: \textit{race}.
In contrast, using \deflev leads to significantly lower scores (F$_1$=0.32).
Lastly, with \defjor Llama3 achieves F$_1$=0.67, falling behind best results, yet still largely outperforming its variant using \deflev.
% 
% 
% We observe that Llama3 achieves the lowest performance with \devlev.
% In contrast, with \defisr and \defsyr, Llama3 achieves the best ones with a F$_1$ score of 0.74.
% \deflev obtains the lowest score, whereas \defisr and \defsyr achieve the highe
% The highest score is obtained by Def. Israel and Def. Syria, with an F$_1$ score of 0.74, which overcomes most of the English results.
% DEF. Syria is the least comprehensive, with only one target component: race.
% 
% \begin{table}[!t]
%     \centering
%     \small
%     \begin{tabularx}{\linewidth}{lXXXX} % 'X' columns will automatically adjust width
%         \toprule
%         \bf Definition & \deflev & \defisr & \defsyr & \defjor \\
%         \midrule
%         Llama3     & 0.32       & 0.74       & 0.74       & 0.67       \\
%         \bottomrule
%     \end{tabularx}
%     \caption{F1 Score on the Positive Class}
%     \label{tab:results_comp}
% \end{table}
% 
We also compute the F$_1$ score across the four definitions. 
Each time, we assume that one definition represents the gold standard when compared to another one.
% and it is compared to another.
Table \ref{tab:results_cross} shows the results. 
The highest F$_1$ (0.75) is reached between \defsyr and \defisr, followed by \deflev and \defisr, with a small 0.01 point difference. 
However, in terms of components, the definitions used in Israel and Jordan are the most similar, as they share three target components: ethnicity, religion, and race.
Table \ref{tab:overlapping} shows the number of overlapping instances across the four definitions, divided into positive (P) and negative (N). 
The agreement between \defsyr and \defisr is confirmed by the number of overlapping instances, as they have the greatest agreement with 478 instances.
In all cases, most of the overlapping instances are on the negative class, as it is the class that the model tends to overpredict.

\begin{table}[!t]
\centering
\small
\begin{tabularx}{\columnwidth}{l|RR|RR|RR}
\toprule
\bf Def.  & \multicolumn{2}{c}{\bf \defisr} & \multicolumn{2}{c}{\bf \defsyr} & \multicolumn{2}{c}{\bf \defjor} \\
         &\bf P & \bf N & \bf P & \bf N & \bf P & \bf N \\
\midrule
\deflev       & 55    & 309   & 55    & 295   & 55    & 339   \\
\defisr       & -     & -     & 190   & 288   & 157   & 299   \\
\defsyr       & -     & -     & -     & -     & 166   & 294   \\
\bottomrule
\end{tabularx}
\caption{Number of overlapping instances in the cross-cultural setting with separate positive (P) and negative (N) values.}
\label{tab:overlapping}
\end{table}

\section{Conclusions}
\label{sec:conclusion}
We introduced \corpus, a comprehensive hate speech definition dataset. 
\corpus provides detailed component annotations, capturing the target, intent/purpose, and act/means of collected hate speech instances that allow comparing hate speech definitions.
Our analysis of \corpus, revealed a lack of cultural diversity in existing definitions. This is primarily because only legislative sources referred to specific cultures. Wikipedia definitions are also often the result of collective contributions or translations of the original English text, making them less culturally distinct. There is also a variation in terms of components,
especially when one language can refer to multiple cultures, such as in the case of Levantine Arabic.
% Moreover, we evaluated how definitions representing different components and cultures affect LLM prompting.
Moreover, our experiments showed that LLMs are sensitive to the employed hate speech definitions, where, in some cases, more comprehensive definitions lead to better results.

Our work
% The \corpus analysis and the LLM experiments
underscores two key considerations for hate speech detection research: (a)~Definitions to be incorporated into the annotation guidelines or prompts must be specific to the task and should clarify the level of comprehensiveness or generality required for that task. This consideration should also align with model selection, as some models perform better with general definitions while others with more comprehensive ones. (b)~Definitions must be relevant to the language and the target culture. This may involve referring to hate speech legislation, to understand what constitutes hate speech in the given culture, otherwise clarify that no culture is considered.

In future work, we plan to develop a community knowledge base of hate speech definitions to foster research in several emerging cultures and domains in this field.

\section*{Limitations}\label{sec:limitations}
Despite our efforts to gather as many definitions as possible for \corpus, finding non-English or culture-specific definitions remains a challenge. As discussed in Section~\ref{sec:hatedefcon_creation}, most of the non-English definitions we obtain are sourced from legislation, while in the case of online platforms and tech companies, they are often translations of the English entries. This limits the breadth of cross-cultural analysis and restricts the scope of our study.

\section*{Acknowledgments}
K. Korre's research is carried out under the project \textit{RACHS: Rilevazione e Analisi Computazionale dell’Hate Speech in rete}, in the framework of the PON programme FSE REACT-EU, Ref.~DOT1303118. 
F. Ruggeri is partially supported by the project European Commission's NextGeneration EU programme, PNRR -- M4C2 -- Investimento 1.3, Partenariato Esteso, PE00000013 - ``FAIR - Future Artificial Intelligence Research'' -- Spoke 8 ``Pervasive AI’’.
Arianna Muti's research is supported by the European Research Council (ERC) under the European Union’s Horizon 2020 research and innovation program (grant agreement No. 101116095, PERSONAE).

\section*{Ethics Statement}\label{sec:ethics_statement}
We clarify that, while some definitions may be more comprehensive or specific than others and acknowledge that definitions can be culture-bound, no definition should be considered inherently superior to another based solely on cultural context. As evident from the literature, definitions often reflect the values of the culture they originate from. The absence of certain components in a definition (such as not including gender as a target) may reflect a more conservative stance in some cases. However, the purpose of this work is not to scrutinize cultures by using hate speech definitions as a proxy, but rather to highlight the fundamental differences that must be considered when applying these definitions for hate speech detection. In other words, the definitions should not only be tailored to the task at hand, but also be specific to the cultural context they represent.

With regard to the annotation, all annotators gave their full consent to participate in this study.

\section*{Working setting}
Our experiments were carried out on an Nvidia A100 GPU, running for 20 minutes on average for each experiment. To limit computational overhead, we did not experiment with more powerful models.

\section*{AI-assisted technologies}
ChatGPT 3.5 has been used to improve the language and readability of the paper. We take full responsibility for the content, which has been properly reviewed and edited to reflect our own methods.

\bibliography{acl_latex}
\bibliographystyle{acl_natbib}

% \newpage

\appendix

\section{Semantic Componential Analysis} \label{sec:appendix:sca-example}

Table~\ref{tab:appendix:kinship} illustrates an example of SCA application.

\begin{table}[!h]
    \centering
    \resizebox{\columnwidth}{!}{
    \begin{tabular}{lccccc}
        \toprule
           & Parent & Sibling & Male & Female & \\
        \midrule
        Father    & + & - & + & - \\
        Mother    & + & - & - & +  \\
        Brother   & - & + & + & -  \\
        Sister    & - & + & - & +  \\
       \bottomrule
    \end{tabular}
    }
    \caption{Kinship terms characterized by attribute dimensions. The attribute dimensions include Parent, Sibling, Male, Female. Each example is marked with '+' if the feature is present, '-' if absent.}
    \label{tab:appendix:kinship}
\end{table}

\section{Laws from the Global Handbook on Hate Speech Laws}
\label{app:laws_list}

We provide the full list of countries included and excluded from \corpus.

\paragraph{Included.} 
Afghanistan, Albania, Algeria, Andorra, Angola, Argentina, Armenia, Australia, Austria, Azerbaijan, Belarus, Belgium, Bolivia, Bosnia and Herzegovina, Botswana, Brazil, Brunei Darussalam, Bulgaria, Cambodia, Cameroon, Canada, Central African Republic, Chad, Chile, China, Colombia, Croatia, Cuba, Cyprus, Czech Republic, Democratic Republic of Congo, Denmark, Djibouti, Estonia, Ethiopia, Fiji, Finland, France, Gabon, Georgia, Germany, Ghana, Greece, Guinea, Guinea-Bissau, Guyana, Haiti, Hungary, Iceland, India, Indonesia, Iran (Islamic Republic of), Iraq, Ireland, Italy, Japan, Jordan, Kenya, Kyrgystan, Latvia, Luxembourg, Myanmar, Malaysia, Malta, Mexico, Moldova, Monaco, Myanmar, Nepal, Netherlands, New Zealand, Norway, Oman, Pakistan, Sweden, Syrian Arab Republic, Tanzania, Timor-Leste, Trinidad and Tobago, Tunisia, Turkmenistan, Uganda, United Arab Emirates, United States of America, Uzbekistan, Venezuela, Zambia, Poland, Portugal, Romania, Russian Federation, Rwanda, Senegal, Serbia, Sierra Leone, Singapore, Slovakia, Somalia, South Africa, South Sudan, Spain, Sri Lanka, Switzerland, Tajikistan, Togo, Turkey, Ukraine, United Kingdom, Uruguay, Vietnam, Zimbabwe.

\paragraph{Excluded.}
Madagascar, Malawi, Maldives, Mali, Marshall Islands, Mauritania, Mauritius, Micronesia (Federated States of), Mongolia, Montenegro, Morocco, Mozambique, Namibia, Nauru, Nicaragua, Niger, Nigeria, North Macedonia, Palau, Panama, Papua New Guinea, Paraguay, Peru, Philippines, Qatar, Republic of Korea, Saint Kitts and Nevis, Saint Lucia, Saint Vincent and the Grenadines, Samoa, San Marino, Sao Tome and Principe, Saudi Arabia, Seychelles, Slovenia, Solomon Islands, Suriname, Thailand, Tonga, Tuvalu, Vanuatu, Yemen.

\section{Annotation Guidelines}
\label{app:annotation_guidelines}

\begin{table}[!t]
    \centering
    \footnotesize
    \begin{tabular}{|p{7.25cm}|}
    \toprule
    \textbf{Read and Understand the Definition.} \\
    Carefully read the provided hate speech definition from the source material. \\
    Ensure you understand the context and content before proceeding with annotations. \\
    \midrule
    \textbf{Identify and Annotate Core Components.} \\
    After reading the definition, review all available components listed in each Excel column. \\
    Annotate each component with 1 if the component exists in the definition. \\
    Annotate with 0 if the component does not exist in the definition. \\
    If a definition is very general, annotate with 1 on the column “General”.
    If you see words/phrases like sexism, or sexist behavior, please annotate by putting one in gender. \\
    \midrule
    \textbf{Undefined Components.} \\
    If you detect a component that has not been included in the columns, please use the column “Undefined Component” to write it down. \\
    \midrule
    \textbf{Add Comments.} \\
    Feel free to add any relevant comments in the comments column. \\
    \bottomrule
    % {\scriptsize
    \textbf{Note:} If a component or a morphological derivative of the component appears in the definition, mark the respective column with a positive annotation. 
    For instance, if the predefined component is “abuse” and the component “abusive” is found in the definition, place a positive annotation (i.e., 1) in the “abuse” column.
    % } 
    \\
    \bottomrule
    \end{tabular}
\caption{\corpus annotation guidelines.}
\label{tab:appendix:guidelines}
\end{table}

Table~\ref{tab:appendix:guidelines} reports annotation guidelines for building \corpus via SCA.

\section{Inter-annotator Agreement for Componential Annotation}
\label{app:iaa}

Table~\ref{tab:appendix:iaa} reports pairwise IAA on \corpus.

\begin{table}[!t]
\centering
\small
\begin{tabularx}{\columnwidth}{Xr}
\toprule
\textbf{Annotator Pair} & \textbf{Average Kappa} \\
\midrule
Annotator$_1$ vs. Annotator$_2$ & 0.75 \\ 
Annotator$_1$ vs. Annotator$_3$ & 0.64 \\ 
Annotator$_2$ vs. Annotator$_3$ & 0.64 \\ 
\bottomrule
\end{tabularx}
\caption{Average Cohen's Kappa scores per annotator pair}
\label{tab:appendix:iaa}
\end{table}

\section{Complete Component Hierarchy and Further Statistics}
\label{app:comp_hierarchy}

Table~\ref{tab:comp_hierarchy} reports the fine-grained SCA hierarchy extracted from \corpus.

\begin{table*}[!t]
\small
\centering
\begin{tabular}{|l|l|}
\hline
\textbf{Hate Speech Framework} & \textbf{Categories} \\ \hline

\multirow{2}{*}{\textbf{Target}} & \textbf{Demographics and Identity} \\
 & \begin{tabular}[c]{@{}l@{}} Race/Ethnicity: Race, Ethnicity, Tribe, Color, Nationality, Regional Origin, \\ Genetic Origin \\
 Nationality/Region: Nationality, Region, Place of Birth, Place of Origin, \\ Place of Residence, Immigration Status \\
 Religion/Belief: Religion, Belief, Creed, Ideology, Philosophical Opinion, \\ Philosophical Ideology, Worldview \\
 Gender/Sexual Orientation: Gender, Sexual Orientation, Gender Identity, \\ Sex Change \\
 Disability: Disability, Physical Condition, Mental Capacity, Health Characteristics, \\ Ability \\
 Age/Appearance: Age, Appearance, Generation \end{tabular} \\ \cline{2-2} 
 & \textbf{Social and Economic Roles} \\
 & \begin{tabular}[c]{@{}l@{}} Occupation/Profession: Occupation, Profession, Job, Employment, Trade \\ Union Membership, Calling \\
 Family Status: Family Status, Marital Status, Familial Status, Pregnancy \\
 Citizenship/Legal Status: Citizenship, Legal Status, Immigration, Veteran Status, \\ Refugees, Nationality \end{tabular} \\ \cline{2-2} 
 & \textbf{Social and Economic Class} \\
 & \begin{tabular}[c]{@{}l@{}} Socioeconomic Status: Economic Status, Social Class, Financial Status, Wealth, \\ Poverty, Social Origin, Social Strata, Economic/Social Origin \\
 Caste/Tribe: Caste, Tribal Affiliation, Ancestry, Descent \end{tabular} \\ \hline

\multirow{3}{*}{\textbf{Intent/Purpose}} & \textbf{Discrimination and Prejudice} \\
 & \begin{tabular}[c]{@{}l@{}} Discrimination: Discrimination, Discriminatory Practices, Exclusion, \\ Marginalization, Segregation, Denigration \\
 Prejudice: Prejudice, Bias, Stereotyping, Xenophobia, Ethnocentrism, Bigotry, \\ Contempt, Superiority \\
 Humiliation: Humiliation, Demeaning, Belittling, Degrading, Ridiculing, Mocking, \\ Stigmatizing, Inferiorizing, Dehumanizing \end{tabular} \\ \cline{2-2} 
 & \textbf{Hostility and Aggression} \\
 & \begin{tabular}[c]{@{}l@{}} Violence: Physical Violence, Aggression, Abuse, Threat, Brutalization, \\ Persecution, Terrorism, Hostility, Rancor \\
 Hate: Hatred, Ill-Will, Animosity, Abhorrence, Detestation, Malice, \\ Anti-Semitism, Racism, Ethnocentricism \\
 Conflict: Conflict, Discord, Dissension, Sectarianism, Division, Social Unrest, \\ Civil Unrest, War \end{tabular} \\ \cline{2-2} 
 & \textbf{Social and Cultural Control} \\
 & \begin{tabular}[c]{@{}l@{}} Cultural Control: Cultural Manipulation, Propaganda, Ideological Imposition, \\ Social Control, Social Hatred, Supremacy, Perpetuation of Norms, \\ Customs, Traditions, Cultural Values \\
 Exclusion: Exclusion, Social Marginalization, Social Exclusion, Economic Exclusion, \\ Stigmatization, Alienation, Isolation \\
 Suppression: Suppression, Silencing, Censorship, Restriction, Limitation of Rights, \\ Deprivation, Harassment \end{tabular} \\ \hline

\multirow{3}{*}{\textbf{Act/Means}} & \textbf{Verbal and Written Expressions} \\
 & \begin{tabular}[c]{@{}l@{}} Insults: Insults, Pejoratives, Slurs, Offensive Language, Derogatory Language, \\ Humiliation, Threat \\
 Defamation: Defamation, Slander, Vilification, Disparagement, Discrediting, \\ Ridicule, Mockery \\
 Provocation: Provocation, Incitement, Inflammatory Speech, Sedition, Antagonism, \\ Triggering, Threats \\
 Misinformation: Misinformation, Disinformation, Propaganda, Deception, Promoting \\ Xenophobia, Racism, and Bigotry \end{tabular} \\ \cline{2-2} 
 & \textbf{Physical Actions} \\
 & \begin{tabular}[c]{@{}l@{}} Violence: Physical Harm, Assault, Attack, Damage to Property, Brutalization, \\ Persecution \\
 Exclusion: Exclusion, Segregation, Denial of Rights, Obstructing Rights, Deprivation, \\ Harassment \\
 Cultural Actions: Desecration, Desecration of Symbols, National Flag Desecration, \\ Denial of Cultural Identity \end{tabular} \\ \cline{2-2} 
 & \textbf{Social and Cultural Manipulation} \\
 & \begin{tabular}[c]{@{}l@{}} Social Control: Manipulation of Social Norms, Cultural Domination, Supremacy, \\ Cultural Stereotyping, Perpetuation of Prejudice \\
 Cultural Exclusion: Cultural Alienation, Cultural Stigmatization, Exclusion from \\ Cultural Activities, Denial of Cultural Identity \\
 Economic Suppression: Economic Marginalization, Social Segregation, Restriction \\ of Economic Opportunities, Denial of Economic Rights \end{tabular} \\ \hline

\end{tabular}
\caption{Hate Speech Framework Hierarchical Structure.}
\label{tab:comp_hierarchy}
\end{table*}

\section{Hate Speech Definitions for GHC} \label{app:ghc-definitions}

Table~\ref{tab:appendix:hate_speech_definitions} reports the definitions used in our experiments with LLMs on GHC.

\begin{table*}[!t]
    \small
    \centering
    \begin{tabular}{c p{0.9\linewidth}   }
        \toprule
        \textbf{No.} & \textbf{Definition}  \\
        \midrule
        \defghc & Language that intends to — through rhetorical devices and contextual references — attack the dignity of a group of people, either through an incitement to violence, encouragement of the incitement to violence, or the incitement to hatred.\\
        \midrule
        \defwiki & Hate speech — a term that denotes speech intended to degrade, disturb, or cause violence or actions based on prejudice against persons or groups of people on the basis of their race, gender, age, ethnicity, nationality, religion, sexual orientation, gender identity, disability, language ability, moral or political views, socioeconomic class, occupation or appearance (such as height, weight, and hair color), mental capacity, and any other characteristic. The term refers to both written and oral communication, as well as some forms of behavior in a public place. Hate speech operates outside the law, speech that offends a particular person or group in terms of discrimination against that person or group. According to the law, hate speech is any speech, gesture or behavior, written text, or display that is prohibited because it is likely to incite violence or prejudice against or by a protected individual or group, or that degrades or intimidates a particular individual or group. The law may recognize the protected individual or group according to certain characteristics.\\
        \midrule
        \defdict & Speech expressing hatred of a particular group of people. \\
        \bottomrule
    \end{tabular}
    \caption{Definitions of Hate Speech. D$_{ghc}$ is the one used to annotate GHC, D$_{wiki}$ is from Wikipedia, and D$_{dict}$ is from the Mariam Webster Dictionary. %Def.2 is the most comprehensive, i.e., it has more components, while Def. 3 is very general and has no component.
    }
    \label{tab:appendix:hate_speech_definitions}
\end{table*}

\section{Definitions for Cross-Cultural Analysis} \label{app:prompts}

We select three definitions from our corpus. 
%Def. 1 is from the GAB research paper, Def.2 is from Wikipedia and Def 3. from the Mariam Webster Dictionary. Def.2 is the most comprehensive, i.e., it has more components, while Def. 3 is very general and has no component. Table
Table~\ref{tab:prompt_def} shows the prompt and the definitions used for the experiments.

\begin{table*}[!t]

\renewcommand{\arraystretch}{1.5}
    \centering
    \footnotesize
    \resizebox{1.0\linewidth}{!}{%
\begin{tabular}{p{2.1\columnwidth}}
\toprule
   \bf Prompt \\
    \midrule
        Read carefully the definition of 'hate speech' provided. 
Your task is to classify the input text as containing hate speech or not. You can only rely on the definition provided.
Respond only with YES or NO. \\ 
Definition: \{\texttt{definition}\} \\
Text: \{\texttt{text}\} \\
Answer: \\
\midrule
\bf Definitions\\
\midrule
\deflev. Hate speech (HS) is formally defined as “any communication that disparages a person or a group on the basis of some characteristic such as race, color, ethnicity, gender, sexual orientation, nationality, religion, or other characteristic” \cite{mulki-etal-2019-l}
\\
\midrule
\defisr. In this Article: “racism” – persecution, humiliation, degradation, a display of enmity, hostility or violence, or causing violence against a public or parts of the population, all because of their colour, racial affiliation or national ethnic origin. Publication of racist incitement is prohibited Article 144B: (a) If a person publishes anything in order to incite to racism, then he is liable to five years imprisonment. (b) For the purposes of this section, it does not matter whether the publication did cause racism, and whether or not it is true. Article 144C: Permissible publication (a) Publication of a true and fair report of an act said in section 144B shall not be deemed an offense under that section, on condition that it was not intended to cause racism. (b) Publication of quotes from religious scriptures or prayer books or the observance of a religious ritual shall not be deemed an offence under section 144B, on condition that it was not intended to cause racism. Article 144D: Possession of racist publication If a person holds a publication prohibited under section 144B for distribution, in order to cause racism, then he is liable to one-year imprisonment, and the publication shall be confiscated. 
\\
\midrule
\defsyr. Criminal Code Article 306: Any act, piece of writing or speech that is intended to or results in stirring sectarian or racial strife or inciting conflict between sects and the various elements of the nation shall be punished by imprisonment of six months to two years and a fine of one hundred to two hundred Syrian pounds, as well as with prohibition from exercising the rights mentioned in the second and fourth paragraphs of Article 65. Article 65: Every person sentenced to imprisonment or house arrest in misdemeanor cases is deprived throughout the execution of his sentence from exercising the following civil rights: A:The right to assume public employment and services. B: The right to assume jobs and services in managing the affairs of the civil sect or managing the union to which he belongs. C: The right to be a voter or elected in all state councils. D: The right to be a voter or elected in all sects and trade union organizations. E: The right to wear Syrian or foreign medals. \\
\midrule
\defjor. Criminal Code Section 5: Crimes Harming National Unity and the Coexistence between the Nation’s Elements Article 150: Any writing or speech aims at or results in stirring sectarian or racial prejudices or the incitement of conflict between different sects or the nation’s elements, such act shall be punished by imprisonment for no less than six months and no more than three years and a fine not to exceed five hundred dinars (JD500). Audiovisual Media Law Article 20(l)(2) prohibits licensed broadcasters from broadcasting hateful, terrorist, violent or seditious material or from promoting religious, sectarian or ethnic strife. \\
\bottomrule
    \end{tabular}}
    \caption{Prompt and definitions used in the cross-cultural experiments.}
    \label{tab:prompt_def}
\end{table*}

\end{document}